\documentclass[entropy,article,accept,moreauthors,pdftex,10pt,a4paper]{mdpi}

\firstpage{1} 
\articlenumber{221}
\doinum{10.3390/e21030221}
\pubvolume{21}
\pubyear{2019}
\copyrightyear{2019}
\history{}
\pdfoutput=1

 \theoremstyle{mdpi}
 \newcounter{thm}
 \newcounter{ex}
 \newcounter{re}
 \theoremstyle{mdpidefinition}

\usepackage{longtable}

\Title{Supervised and Unsupervised End-to-End Deep Learning for Gene Ontology Classification of Neural In-Situ Hybridization Images$^\S$}

\Author{Ido Cohen *, Eli (Omid) David * and Nathan S. Netanyahu *}
\AuthorNames{Ido Cohen, Eli (Omid) David, and Nathan S. Netanyahu}

\address{Department of Computer Science, Bar-Ilan University, Ramat-Gan 5290002, Israel}

\corres{Correspondence: cido15@gmail.com (I.C.); mail@elidavid.com (E.D.); nathan@cs.biu.ac.il (N.S.N.)}

\abstract{
In recent years large datasets of high-resolution mammalian neural images have become available, which has prompted active research on the analysis of gene expression data. Traditional image processing methods are typically applied for learning functional representations of genes, based on their expressions in these brain images. In this paper, we describe a novel end-to-end deep learning-based method for generating compact representations of \emph{in situ hybridization} (ISH) images, that are invariant-to-translation. In contrast to traditional image processing methods, our method relies, instead, on deep \emph{convolutional denoising autoencoders} (CDAE) for processing raw pixel inputs, and generating the desired compact image representations.
We provide an in-depth description of our deep learning-based approach, and present extensive experimental results, demonstrating that representations extracted by CDAE can help learn features of functional \emph{gene ontology categories} for their classification in a highly accurate manner. Our methods improves the previous state-of-the-art classification rate~\cite{Liscovitch2013FuncISH} from an average AUC of 0.92 to 0.997, i.e., it achieves 96\% reduction in error rate. Furthermore, the representation vectors generated due to our method are more compact in comparison to previous state-of-the-art methods, allowing for a more efficient high-level representation of images. These results are obtained with significantly downsampled images in comparison to the original high-resolution ones, further underscoring the robustness of our proposed method.
}


\conference{\cite{Cohen2017deepbrain}}

\usepackage[overlay,absolute]{textpos}

\begin{document}

\begin{textblock*}{10in}(58mm, 10mm)
{\textbf{Ref:} \emph{Entropy}, Vol.~21, No.~3, pp.~221--238, February 2019.}
\end{textblock*}

\section{Introduction}

A very large volume of high-spatial resolution imaging datasets is available these days in various domains, calling for a wide range of exploration methods based on image processing. One such dataset has become recently available in the field of neuroscience, thanks to the Allen Institute for Brain Science. This dataset contains \textit{in situ hybridization} (ISH) images of mammalian brains, in unprecedented amounts, which has motivated new research efforts \cite{Henry2012atlases}, \cite{Lein2007Genome}, \cite{Ng2009expression}.
ISH is a powerful technique for localizing specific nucleic acid targets within fixed tissues and cells; it provides an effective approach for obtaining temporal and spatial information about gene expression \cite{Puniyani2013GINI}. Images now reveal highly complex patterns of gene expression varying on multiple scales.

Developing analytical tools for discovering gene interactions from such data remains an open challenge due to various reasons, including difficulties in extracting canonical representations of gene activities from images, and inferring statistically meaningful networks from such representations. The challenge in analyzing these images is both in extracting the patterns that are most relevant functionally, and in providing a meaningful representation that would allow neuroscientists to interpret the extracted patterns.

One of the aims at finding a meaningful representation for such images, is to carry out classification to \textit{gene ontology} (GO) categories. GO is a major bioinformatics initiative to unify the representation of gene and gene product attributes across all species \cite{GO2008project}. More specifically, it aims at maintaining and developing a controlled vocabulary of gene and gene product attributes and at annotating them. This task is far from done; in fact, several gene and gene product functions of many organisms have yet to be discovered and annotated \cite{Ashburner2000GO}. Gene function annotations, which are associations between a gene and a term of controlled vocabulary describing gene functional features, are of paramount importance in modern biology. They are used to design novel biological experiments and interpret their results. Since gene validation through in vitro biomolecular experiments is costly and tedious, deriving new computational methods and software for predicting and prioritizing new biomolecular annotations, would make an important contribution to the field \cite{Perez2004annotation}. In other words, deriving an effective computational procedure that predicts reliably likely annotations, and thus speed up the discovery of new gene annotations, would be very useful \cite{Plessis2011why}.

Past methods for analyzing brain images need to reference a brain atlas, and are based on smooth non-linear transformations \cite{Davis2009Allen}, \cite{Hawrylycz2011Multi}. These types of analyses may be insensitive to fine local patterns, like those found in the layered structure of the cerebellum\footnote{The cerebellum is a region of the brain. It plays an important role in motor control, and has some effect on cognitive functions \cite{Wolf2009cerebellar}.}, or to a spatial distribution. In addition, most machine vision approaches address the challenge of providing human interpretable analysis. Conversely, in bioimaging usually the goal is to reveal features and structures that are hardly seen even by human experts. For example, one of the new functions that follow this approach is presented in \cite{Liscovitch2013FuncISH}, using a histogram of local \textit{scale-invariant feature transform} (SIFT)~\cite{Lowe2004SIFT} descriptors on several scales.

Recently, several machine learning algorithms have been designed and implemented to predict GO annotations \cite{Pinoli2015Computational}, \cite{Zitnik2014mold}, \cite{Vembu2014prediction}, \cite{Kordmahalleh2013Hierarchical}, \cite{King2013patterns}.
In our research, we examine an \textit{artificial neural network} (ANN) with many layers (also known as \textit{deep learning}), in order to achieve functional representations of neural ISH images.

To obtain a compact feature representation of these ISH images, we explore, in this paper, \textit{autoencoders} (AE) and \textit{convolutional neural networks} (CNN). Specifically, we demonstrate that the \textit{convolutional autoencoder} (CAE) is most appropriate for the task at hand. (This is similar to the work of Krizhevsky and Hinton \cite{Krizhevsky2011DeepAE}, who used deep autoencoders to create short binary codes for content-based images.) Subsequently, we use this representation to learn features of functional GO categories for every image, invoking a simple $L_2$-regularized logistic regression classifier, as in \cite{Liscovitch2013FuncISH}. As a result, each image is represented as a lower-dimensional vector whose components correspond to meaningful functional annotations.
As pointed out in~\cite{Liscovitch2013FuncISH}, the resulting representations define similarities between ISH images which could be explained, hopefully, by such functional categories.

Our experimental results further demonstrate that the representation obtained using the novel architecture of a so-called \textit{convolutional denoising autoencoder} (CDAE) outperforms the previous state-of-the-art classification rate\cite{Liscovitch2013FuncISH}; specifically, it improves the average AUC from 0.92 to 0.997, achieving 96\% reduction in error. The method operates on input images that are downsampled significantly with respect to the original ones to make it computationally feasible.


\section{Background}
\subsection{Biological Background}
\label{sec:Biological}

\begin{figure}[ht] 
\centering
\includegraphics[width=\textwidth]{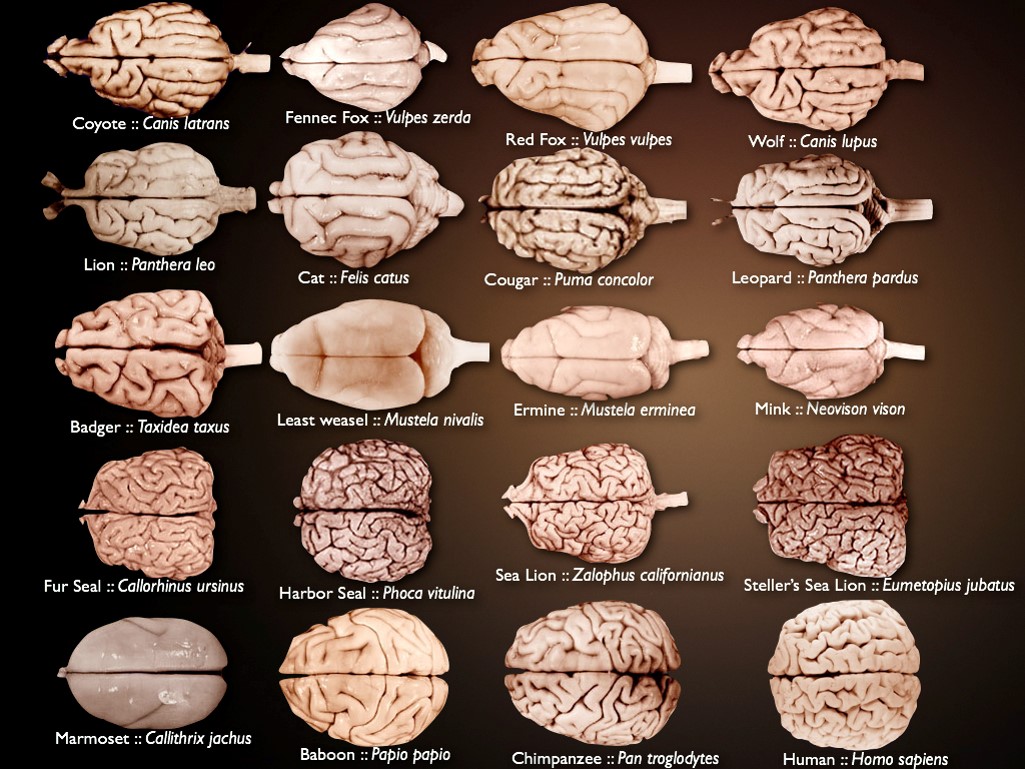}
\caption{Different mammalian brains, varying in size and complexity (source: University of Wisconsin and Michigan State Comparative Mammalian Brain Collections).}

\label{fig:mammalian_brain}
\end{figure}

Mammalian brains vary in size and complexity (Figure \ref{fig:mammalian_brain}) and are composed of billions of neurons and glia. The brain is organized in highly complex anatomical structures.
This is why for many years, it has remained a fascinating challenge.

\begin{figure}[ht] 
\centering
\includegraphics[width=\textwidth]{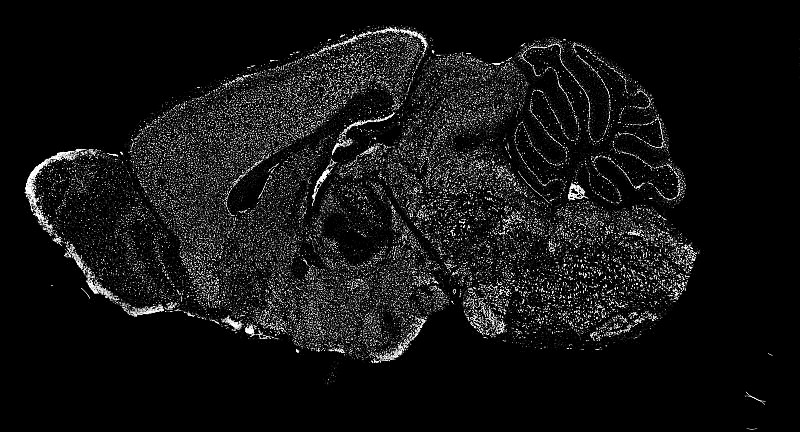}
\caption{Typical in situ hybridization (ISH) image of a mammalian brain used for our research (in gray scale).}
\label{fig:ISH}
\end{figure}

The dataset of the Allen Institute for Brain Science contains a significant repository of ISH images of mammalian brains for many research fields.
ISH is a powerful technique for localizing specific nucleic acid targets within fixed tissues and cells. It provides an effective approach for obtaining temporal and spatial information about gene expression \cite{Puniyani2013GINI}. 
ISH images reveal highly complex patterns of gene expression varying on multiple scales, as shown in Figure~\ref{fig:ISH}. This information calls for a wide range of exploration methods based on image processing.

\begin{figure}[ht] 
\centering
\includegraphics[width=\textwidth]{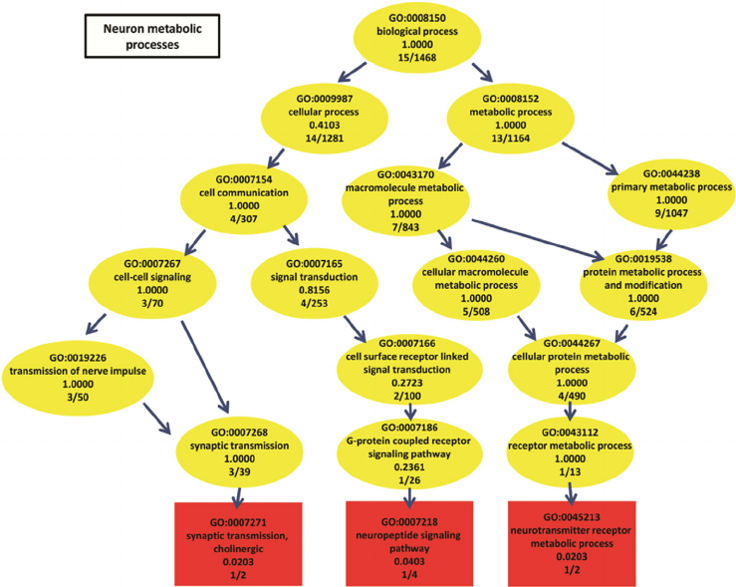}
\caption{Directed acyclic graph representing Neuron metabolic processes (source: The European Bioinformatics Institute (EMBL--EBI)).}
\label{fig:GO_DAG}
\end{figure}

Finding meaningful representations for such images is one of the main goals which could be used for classification of \textit{gene ontology} (GO) categories. GO has the aim of unifying the representation of gene and gene product\footnote{A gene product is the biochemical material, either RNA or protein, resulting from expression of a gene.} attributes across all species \cite{GO2008project}. GO categories are shaped in the form of a \textit{directed acyclic graph} (DAG), in which each category has relevant subcategories, and vice versa. An example can be seen in Figure \ref{fig:GO_DAG}.

Thousands of GO categories exist, and the task is far from done; in fact, several gene and gene product functions of many organisms have yet to be discovered and annotated \cite{Ashburner2000GO}. Gene function annotations, which are associations between a gene and a term of controlled vocabulary describing gene functional features, are of paramount importance in modern biology. They are used to design novel biological experiments and interpret their results. 
With that said, gene validation through in vitro biomolecular experiments is costly and lengthy. It is a multi-stage process, which starts by revealing the gene product and then searching for it in known databases. If it appears in the database, progress is possible; otherwise , a series of laboratory studies are needed to reconstruct the gene. Sometimes it is a short process, but it could also be extremely tedious. For this reason, deriving new computational methods and software for predicting and prioritizing new biomolecular annotations, would make an important contribution to the field~\cite{Perez2004annotation}. In other words, deriving an effective computational procedure that predicts reliably likely annotations, and thus speeding up the discovery of new gene annotations, would be very useful~\cite{Plessis2011why}.

\subsection{FuncISH: Learning Functional Representations} \label{sec:FuncISH}

\begin{figure}[ht] 
\centering
\includegraphics[width=\textwidth]{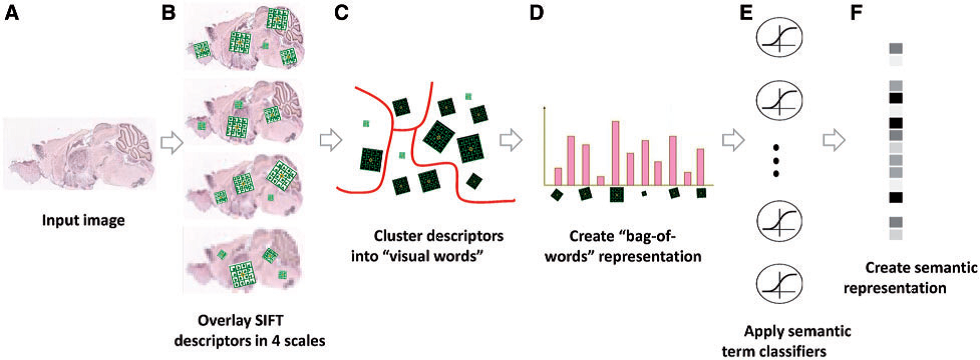}
\caption{Illustration of the image processing pipeline: (a) Original image in pixel grayscale indicating level of gene expression, (b) local SIFT descriptors
are extracted from image at 4 resolutions, (c) descriptors from all 16,351 images are clustered into 500 representative ``visual words'' for each resolution
level using $k$-means, (d) each image is represented as a histogram counting the occurrences of visual words, (e) $L_2$-regularized logistic regression
classifiers applied to 2,081 GO categories, and (f) final 2,081 dimensional image representation (source: Liscovitch \textit{et al}. \cite{Liscovitch2013FuncISH}).}

\label{fig:FuncISH}
\end{figure}

ISH images of mammalian brains reveal highly complex patterns of gene expression varying on multiple scales.
In \cite{Liscovitch2013FuncISH} the authors present \textit{FuncISH}, a learning method of functional representations of ISH images, using a histogram of local descriptors on several scales. They first represent each image as a collection of local descriptors of SIFT features~\cite{Lowe2003panoramas}. Next, they construct a standard \textit{bag-of-words} description of each image, giving a 2004-dimensional representation vector for each gene. Finally, given a set of predefined GO annotations of each gene, they train a separate classifier for each known biological category, using the SIFT bag-of-words representation as an input vector. Specifically, they construct a set of 2,081 \textit{$L_2$-regularized logistic regression} classifiers. GO categories are picked with a number of annotations ranging from 15 to 500 genes. The lower limit is picked to provide enough positive examples for testing, while the higher limit is chosen to preclude the resulting semantic explanations from being too general. 
A scheme taken from \cite{Liscovitch2013FuncISH} representing the work flow is presented in Figure \ref{fig:FuncISH} and Figure \ref{fig:flow_chart_sift}. 

Applying \cite{Liscovitch2013FuncISH} to the genomic set of neural mouse ISH images (available from the Allen Brain Atlas) reveals that ``most neural biological processes could be inferred from spatial expression patterns with high accuracy''. Despite ignoring important global location information, $\sim$700 functional annotations were successfully inferred, and then used to detect gene-gene similarities not captured by previous, global correlation-based methods. According to \cite{Liscovitch2013FuncISH}, combining local and global patterns of expression is an important topic for further research, e.g., the use of more sophisticated non-linear classifiers.

Pursuing further the above classification problem poses a number of challenges. First, defining a certain set of rules that an ISH image has to conform to in order to classify it to the correct GO category. Conventional computer vision techniques, although capable of identifying shapes and objects in an image, seem unlikely to provide effective solutions to the problem of interest. Thus, following the work in~\cite{Liscovitch2013FuncISH}, we use deep learning to achieve more accurate results, due to the more effective functional feature representation of the ISH images learned by our neural network.

\subsection{Convolutional Neural Networks (CNNs)} \label{sec:Convolutional-Neural-Networks}

CNNs are variations of \textit{multilayer perceptrons}, designed to process data that come in the form of multiple arrays, e.g.,~\cite{LeCun1989handwritten},
~\cite{LeCun1998gradient},~\cite{Krizhevsky2012ImageNet}.
1D, 2D, and 3D CNNs are used typically for processing signals/sequences, images/spectrograms, and video/volumetric images, respectively. There are four main components associated with CNNs, all of which exploit the properties of natural signals; these are local connections, shared weights, pooling, and the use of many layers.

\begin{figure}[ht]
	\centering
	\includegraphics[width=\textwidth]{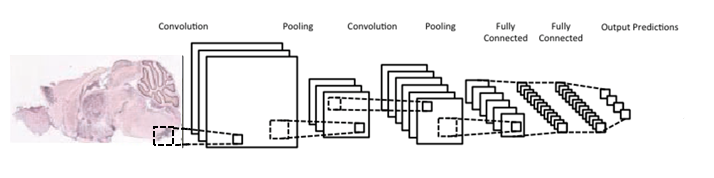}
	\caption{Typical architecture of a convolutional neural network; consists of two convolutional layers (each immediately preceded by a pooling layer), followed by two fully-connected layers, with a final classification layer.}
	\label{fig:my_cnn_arc}

\end{figure}

A typical CNN architecture (Figure \ref{fig:my_cnn_arc}) is structured as a series of components from two types, convolutional layers and pooling layers. Overall, units in a convolutional layer are organized in feature maps, where each unit is connected to a local region of units in the feature maps of the previous layer, through a set of weights called a filter. This local weighted sum is then passed through an activation function, such as the previously defined ReLU(.) or tanh(.) functions, which allows such networks to solve nontrivial problems. All units in a feature map share the same filter weights, while different feature maps in a layer use a different filter. The mathematical advantage is that the number of weights to be learned is not dependent on the number of units, but on the number of feature maps and the size of the filter. This is reasonable because in spatial data such as images, often local values are highly correlated, due to the fact that image  statistics are invariant to location. In other words, if a pattern appears in one part of the image, it could appear anywhere in the image. 

While the role of the convolutional layer is to detect locally connected features from the previous layer, the role of the pooling layer is to merge semantically similar features into one. The pooling layers subsample the data, thus reducing the representation size. For example, a typical pooling method, called max-pooling, selects only the maximum value of a local region of units in one (or more) feature maps (e.g., selects only the maximum value for each $2 \times 2$ region, thus decreasing the size by a factor of 4).
Usually, two or three stages of convolution (using nonlinear functions) and pooling layers are stacked, followed by some fully-connected layers and a final classification layer. Designing a CNN to cope with the relative large size of our input images might be the answer, although it does not provide a solution to our second issue, namely the small number of training samples for each category. 

\subsection{Auto-Encoders (AEs)} \label{sec:Auto-Encoders}

CNNs are effective in a supervised framework~\cite{LeCun1998gradient},~\cite{Krizhevsky2012ImageNet},~\cite{Krizhevsky2011DeepAE},~\cite{Behnke2003hierarchical},~\cite{Krizhevsky2009thesis}, provided a large training set is available.
If only a small number of training samples is available, unsupervised pre-training methods, such as \textit{restricted Boltzmann machines} (RBM) \cite{Hinton2006dbn} or autoencoders \cite{Vincent2008dae}, have proven highly effective.

An AE is a neural network which sets the target values (of the output layer) to be equal to those of the input, using hidden layers of smaller and smaller size, which comprise a bottleneck. Thus, an AE can be trained in an unsupervised manner, forcing the network to learn a higher-level representation of the input. 
One or more hidden layers of neurons are used between the input and output layers, where each layer is usually set to have fewer neurons than those in the input and output layers, thereby creating a bottleneck. This is done with the intention of forcing the network to learn a higher level representation of the input. In this form, each input is first mapped to a hidden layer (smaller than the input layer), and then the output layer tries to reconstruct the original input. Note that for one hidden layer, the set of weights \(w\) between the input layer and the hidden layer, called the "encoder layer", and the set of weights \(w'\) between the hidden layer and the output layer, called the "decoder layer", can be tied (by setting \(w'=w^T\)). Similarly, an AE with more than one hidden layer can maintain tied weights\footnote{A typical equation for a feedforward neural network is \(f(x)=\sigma_2 (b_2+w_2 \sigma_1 (b_1+w_1 x))\) while in autoencoders with tied weights the equation used is \(f(x)=\sigma_2 (b_2+w_1^T \sigma_1 (b_1+w_1 x))\). Setting \(w_2=w_1^T\)  eliminates many degrees of freedom, which has its important advantages.} between the corresponding encoder and decoder layers. In any case, AEs are typically trained using backpropagation with stochastic gradient descent to reduce the reconstruction error \cite{Rumelhart1986errors}, \cite{Werbos1974prediction}.

\begin{figure}[ht]
	\centering
	\includegraphics[]{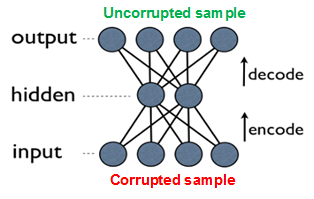}
	\caption{Denoising autoencoder with one layer; during training the input sample is corrupted, while the expected output is the original, uncorrupted input.}
	\label{fig:DAE}
\end{figure}

An improved approach, which outperforms basic autoencoders in many tasks is due to \textit{denoising autoencoders} (DAEs) \cite{Vincent2008dae}, \cite{Vincent2010SDAE}. These are built as regular AEs, where each input is corrupted by added noise, or by setting to zero some portion of the values. Although the input sample is corrupted, the network's objective is to produce the original (uncorrupted) values in the output layer (see Figure \ref{fig:DAE}). Forcing the network to recreate the uncorrupted values results in reduced network overfitting (also due to the fact that the network rarely receives the same input twice), and in extraction of more high-level features. Note that the noise is added only during training, while in prediction time the network is given the uncorrupted input.

For any autoencoder-based approach, once training is complete, the decoder layer(s) are removed, such that a given input passes through the network and yields a high-level representation of the data. In most implementations (such as ours), these representations can then be used for supervised classification.

In the next Section we present our convolutional autoencoder approach, which operates solely on raw pixel data. This supports our main goal, i.e., learning representations of given brain images to extract useful information, more easily, when building classifiers or other predictors. The representations obtained are vectors which can be used to solve a variety of problems, e.g., the problem of GO classification. 
For this reason, a good representation is also one that is useful as input to a supervised predictor, as it allows us to build classifiers for the biological categories known.

\subsection{Convolutional Autoencoders (CAEs)} \label{sec:Convolutional-Autoencoders}

CNNs and AEs can be combined to produce a CAE, which, as mentioned in \cite{Turchenko2015Caffe}, is one of the most preferred architectures in deep learning. 
This may explain why several approaches involving the combination of these methods have been explored previously.
One problem with AEs and DAEs is that both do not account for the local 2D image structure and thus may not be translational-invariant.
Furthermore, when using fully-connected AEs, realistic input sizes may introduce redundancy in the parameters to be learned and increase significantly their number. However, CNNs discover repetitive localized features all over the input, such as in successful models like \cite{Lowe2004recognition}, \cite{Serre2005cortex}. 
As with CNNs, the CAE weights are shared among all locations in the input, preserving spatial locality and reducing the number of parameters.
A common convention for combining CNNs with AEs (or DAEs), which maintains the input size, is to have for each encoder layer a corresponding decoder layer, i.e., each convolutional layer would have a \textit{deconvolutional} layer, and each max-pooling layer would have an \textit{unpooling} layer, that restores the original values lost by the max-pooling subsampling.

Deconvolutional layers are essentially the same as convolutional layers, and similarly to standard autoencoders, they can either be learned or set equal to (the transpose of) the original convolutional layers, as with tied weights in autoencoders.

\begin{figure}[ht]
	\centering
	\includegraphics[width=\textwidth]{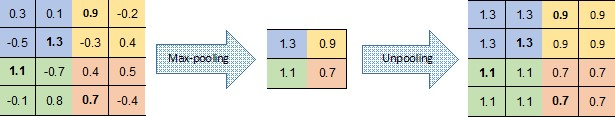}
	\caption{Pooling and unpooling layers; for each pooling layer, the max value is kept, and then duplicated in the unpooling layer.}
	\label{fig:pooling}
\end{figure}

There are a number of variants to the \textit{unpooling} operation (see, e.g., \cite{zeiler2014Visualizing}, \cite{masci2011scae} ,and \cite{Zeiler2011adaptive}). In our CAE, the $2 \times 2$ unpooling simply stores the next layer (Figure ~\ref{fig:pooling}).
This means that there is no need to record the locations of the maxima within each pooling region.

\begin{figure}[ht]
	\centering
	\includegraphics[width=\textwidth]{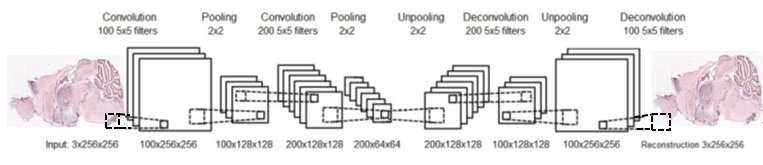}
	\caption{Illustration of convolutional autoencoder; CAE consists of two convolutional layers and their two corresponding deconvolutional layers, and two max-pooling layers and their corresponding unpooling layers.}
	\label{fig:cae_arc}
\end{figure}

Similarly to an AE, after training a CAE, the unpooling and deconvolutional layers are removed. At this point, a neural net, composed from convolution and pooling layers, can be used to find a functional representation, as in our case, or initialize a supervised CNN.
An example of a complete structure of a convolutional autoencoder is presented in Figure~\ref{fig:cae_arc}. 

Similarly to a DAE, a CAE with input corrupted by added noise is called a \textit{convolution denoising autoencoders} (CDAE).

\subsection{Challenges of Applying End-to-End Deep Learning for Annotating Gene Expression Patterns}

Applying end-to-end deep learning for functional representation is especially challenging from an information-theoretic point of view. In this subsection we analyze the challenges present and methods for overcoming them.

In the traditional approach to problems of this nature, raw data (e.g., pixels) are not fed directly into the training module, and instead, a feature extraction phase is first performed, where raw data are converted into a list of features. When applying end-to-end deep learning, however, raw data are directly fed into the input layer of a deep neural network. The idea is that the neural network would perform \emph{implicit feature learning}, and learn more complex and meaningful nonlinear features that would be superior to the handcrafted features by human experts.

While the results obtained due to deep learning represent major improvements over the traditional approach in nearly every domain, deep learning requires substantially larger amounts of training data. While in the traditional approach the model already receives useful features, and needs only to learn how to use those features for successful prediction, in deep learning the features themselves need to be learned as well.

In the case of gene annotation of expression patterns, we only have thousands of images for training, which are not sufficient, by themselves, for obtaining optimal results. One successful approach in such a case is to use \emph{transfer learning}, i.e., first train the neural network on a different dataset (for which ample data are available), and then use that pretrained neural network for training on the task at hand. The idea is that the neural network would learn to perform feature extraction by training it on the larger dataset, and then fine-tune the feature set for the desired task. This would work well assuming that the larger dataset and the target dataset have a similar sample distribution.
(i.e., concepts learned from the larger dataset are transferable to the other).

Zeng \emph{et al.}~\cite{Zeng2015dcnn} employ such a transfer learning approach; in order to obtain a CNN to annotate, eventually, gene expression patterns, they train a model from \emph{OverFeat}~\cite{Sermanet2014overfeat}
with natural images during the pretraining phase, and then using this pretrained network (using natural images) as feature extractors on ISH images. This follows recent studies, using \emph{ImageNet} data \cite{Krizhevsky2012ImageNet} (an image dataset with thousands of categories and millions of labeled natural images), to train a CNN model. The trained CNN is then used as a feature extractor for other datasets, yielding promising performance.

Using the OverFeat model, Zeng \emph{et al.} first resize all images to $231 \times 231$, as required by the model. They then pass each ISH image through the CNN, extracting feature vectors only from the last layer in each stage. These vectors are flattened and concatenated into a single feature vector for each layer. Finally, to build the representation vector, they extract features for each section separately and compute an element-wide maximum across feature vectors from all sections of the same experiment of a gene expression. This global feature vector of size 10,521 represents the gene expression of a given ISH image. This representation is then used to perform gene expression pattern annotation, similarly to \cite{Liscovitch2013FuncISH}, i.e., by training on $L_2$-regularized logistic regression classifiers. The overall average AUC obtained was $0.894 \pm 0.014$, compared with $0.820 \pm 0.046$, due to a similar bag-of-words approach presented in \cite{Liscovitch2013FuncISH}.

Even though the results of Zeng \emph{et al.} show some improvement obtained by transfer learning, these gains are limited, since the ImageNet dataset they have used for pretraining contains natural images, which are vastly different from images intended for gene annotation. In other words, pretraining a neural network on real-world natural images might not be entirely beneficial for its subsequent training for the gene attribution task in question.

In the next section we present our novel method, which uses pretraining on the same sample distribution (as in~\cite{Liscovitch2013FuncISH}), via our proposed unsupervised learning. This pretraining feature learning will prove more effective for the desired task, in the terms of the eventual accuracy gained.


\section{Proposed Approach}
\label{sec:research}

We now discuss our approach for generating a functional representation of ISH images. Deep learning methods have proven recently very successful in finding compact feature representations for many tasks. These networks are trained extensively, in an attempt to capture object representations that are invariant to changes in location, and to some extent also to scale, orientation, lighting, etc.
CNNs and AEs, as described in the Background, achieve small errors when used in a supervised and unsupervised learning modes, respectively. Challenging tasks tackled in a supervised learning mode are, e.g., ImageNet \cite{Krizhevsky2012ImageNet} and large-scale image recognition \cite{Simonyan2014scale}; in an unsupervised mode, such tasks include, e.g., mimicking faithfully biological measurements of the visual areas \cite{Lee2007sparse}, building class-specific feature detectors from unlabeled images \cite{Le2012building}, or creating short binary codes for images based on their content, using deep autoencoders according to Krizhevsky and Hinton \cite{Krizhevsky2011DeepAE}.

\begin{figure}[ht]
	\centering
	\includegraphics[width=0.75\textwidth]{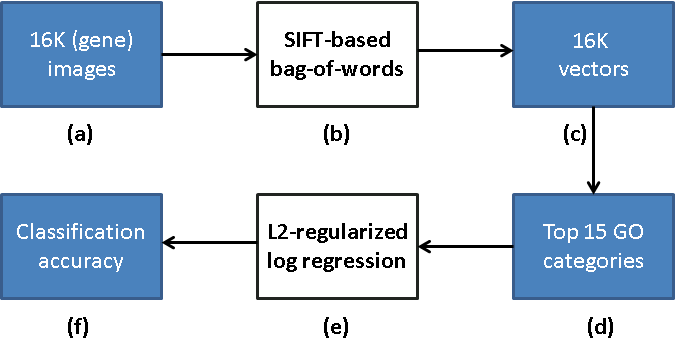}
	\caption{Processing pipeline \cite{Liscovitch2013FuncISH}: (a) 16K original images in pixel grayscale indicating level of gene expression, (b) “bag-of-words” via clustering clustering of SIFT feature descriptors, (c) vector representation due to bag-of-words, (d) 16K vectors trained with respect to each of 15 GO categories with best AUC classification in~\cite{Liscovitch2013FuncISH}, (e) $L_2$-regularized logistic regression classifiers for top 15 GO categories, and (f) classification accuracy obtained.}
	\label{fig:flow_chart_sift}
\end{figure}

Our research follows \cite{Liscovitch2013FuncISH}, but with a modified approach for learning the gene representation and training the GO classifiers. We use CNNs and CAEs to learn the representation, and then use the same classification method as in \cite{Liscovitch2013FuncISH} to measure the classification accuracy (that obviously depends on our different representation). The scheme representing the work flow of \cite{Liscovitch2013FuncISH} is presented in Figure~\ref{fig:flow_chart_sift}, while our work flow is presented in Figure~\ref{fig:flow} (see Subsection~\ref{sec:CDAE-classification}). 

We applied our method to the genomic set of neural mouse ISH images available from the Allen Brain Atlas. This dataset contains whole-brain, expression-masked images of gene expression measured using ISH.
For each gene, a different adult mouse brain was sliced into 25 slices, each 100[mm] thick. We used the most medial slice for each series. In this way, the expression was measured for the entire mouse genome, for a total of 15K genes. That is, the dataset includes 16,351 images representing 15,612 genes. Each image reveals highly-complex patterns of gene expression varying on multiple scales.

To learn detectors of functional GO categories, using the representation obtained for every gene, we introduce a novel CDAE architecture for the problem in question, to find the compact representation of each ISH image. The CDAE includes pooling and unpooling layers, as well as convolution and deconvolutional layers, as was explained in the Background. The input is corrupted for denoising purposes, and should be normalized, in principle, by subtracting its mean from each pixel value and dividing by its standard deviation. As for any AE, layers are trained in an unsupervised manner, using backpropagation for updating the network's weights; this is a well-known procedure for feature extraction to gain a compact representation.
After the CDAE is trained, the deconvolution and unpooling layers are removed, so that the middle (and smallest) hidden layer becomes the output layer. Feeding this network a single ISH image at a time, the resulting output serves as the compact representation of this image, i.e., the representation of the corresponding gene.
These representations are vectors which can then be used to solve a variety of problems, among others, the problem of GO classification.

\subsection{CDAE for GO Classification} 
\label{sec:CDAE-classification}

\begin{figure}[ht!]
	\centering
    \includegraphics[width=0.75\textwidth]{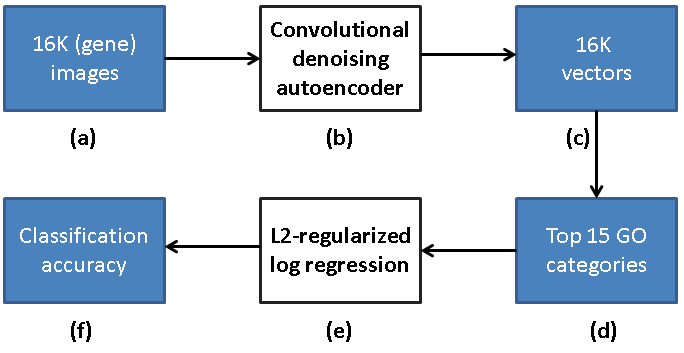}
	\caption{Our work flow: (a) 16K grayscale images indicating level of gene expression, (b) CDAE-based feature extraction for a compact vector representation of each gene, (c) vector representation due to feature extraction, (d) 16K vectors trained with respect to each of 15 GO categories with best AUC classification in \cite{Liscovitch2013FuncISH}, (e) $L_2$-regularized logistic regression classifiers for top 15 GO categories, and (f) classification accuracy obtained.}
	\label{fig:flow}
\end{figure}

Figure~\ref{fig:flow} depicts a framework for capturing the representation similar to FuncISH. Recall that a SIFT-based module was used in \cite{Liscovitch2013FuncISH} for feature extraction. Our scheme learns, alternatively, a CDAE-based representation, before carrying out the GO classification.

For the unsupervised training of our CDAE we use the genomic set of neural mouse ISH images available from the Allen Brain Atlas, which includes 16,351 images representing 15,612 genes. These JPEG images have an average resolution of $16,000 \times 8,000$ pixels. To obtain a representation vector of size $\sim$2,000, the images were resampled initially to $960 \times 480$, and were eventually downsampled (via successive max unpooling operations) to $60 \times 30$.
Despite this heavy down-sampling, we believe that local and global patterns of expression are still preserved to a sufficient extent, in the sense that significant differences between ISH images of different genes can still be captured, thus rendering their representation at the above spatial resolution as meaningful. This working assumption is supported by the down-sampling done also  in~\cite{Liscovitch2013FuncISH}, which succeeded in generating significant representation, as well as by our own results, as detailed below.

Regarding the supervised GO classification, we used the $L_2$-regularized logistic regression classifier as described in ~\cite{Liscovitch2013FuncISH}; specifically, a two-phase \textit{5-fold cross validation} was invoked, per each category, for training the classifier and tuning the logistic regression regularization hyper-parameter. Training requires careful consideration, in this case, due to the highly-imbalanced nature of the training sets.
While in~\cite{Liscovitch2013FuncISH} the full set of 16K gene images was split into five non-overlapping equal sets (without controlling the number of ``positives'' in each split), we split the ``positives'' into five non-overlapping equal subsets and the ``negatives'' into five non-overlapping equal subsets, and combined a different positive subset with each negative subset to obtain five non-overlapping equal-sized subsets, per each category. We then trained the classifiers on four of these subsets, and tested the performance on the fifth. This procedure was repeated five times, each time with a different test subset.

\section{Empirical Results}
\label{sec:experiments}

\subsection{Determination of Hyper-parameters}
In this section we discuss the determination of various hyper-parameters for training, and present a comprehensive set of experimental results. In principle, the hyper-parameters were chosen by running the system, each time, on a large range of possible values for a given hyper-parameter (while keeping the others fixed), and looking for those values which provide the best results. Using insights gained during this extensive experimentation, allowed us to converge faster, in some cases, to optimal hyper-parameter values, without having to run the system on all possible values.

\subsubsection{Denoising Rate}
\label{denoising_rate}
As described in Subsection \ref{sec:Auto-Encoders}, the idea behind denoising autoencoder (DAE), in general, and our convolution denoising autoencoder (CDAE), in particular, is to force the hidden layer to capture more robust features, and prevent it from simply learning the identity. This is achieved by training the CDAE to reconstruct the input from a corrupted version of it.

\begin{figure}[ht!]
	\centering
	\includegraphics[width=\textwidth]{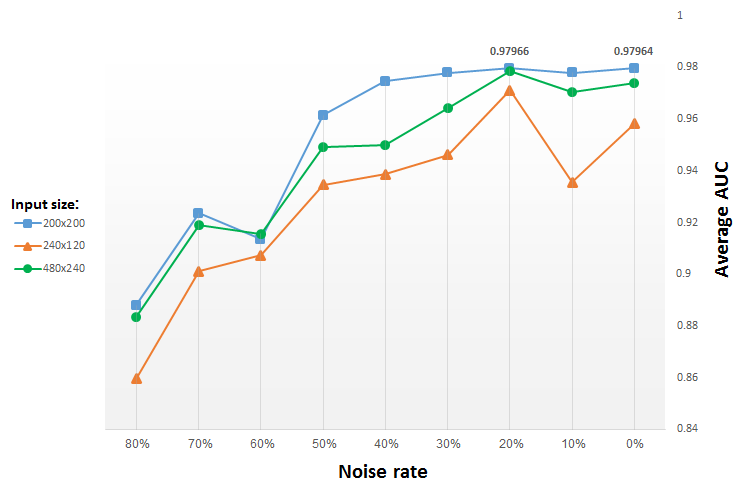}
	\caption{Average AUC for classifiers of top 15 GO categories versus denoising rate (i.e., percentage of pixels randomly set to zero), using CDAE functional representation; each plot corresponds to a different input image size, with other parameters fixed.}
	\label{fig:Denoising_rate}
\end{figure}

We use the same stochastic corruption process as in~\cite{Vincent2008dae}, which randomly sets a predefined amount of the image pixels to zero. Hence the CDAE is forced to predict the missing values from the uncorrupted values, by reconstructing them for randomly selected subsets of missing patterns. Figure \ref{fig:Denoising_rate} presents test results versus the percentage (from 0\% to 80\%) of pixels randomly set to zero, for input images of different size. The blue plot in the figure corresponds to input images resized to $200 \times 200$, yielding (after two max-pooling layers) a compact, ($50 \times 50 =$) 2,500-dimensional vector representation. The green and orange plots both correspond to a compact, $60 \times $30 (vector) representation, but they originate from input images resized to $480 \times $240 and $240 \times $120, respectively. Note that in order to reach the same vector dimensionality, we use an additional pooling layer (and a corresponding unpooling layer) in the network that inputs
$480 \times $240 images.

For each input size, the average AUC drops when the denoising rate increases above 30\% (i.e., when 30\% pixels selected randomly are set to zero). For small denoising rates, the average results improve slightly for added noise, reaching a peak for a denoising rate of roughly 20\%.
Thus we use for most of our study a denoising rate of 20\%, for which the best results are obtained.

\subsubsection{Filter Size}
\label{sec:filter_size}

As mentioned in Subsection \ref{sec:Convolutional-Neural-Networks}, a typical CNN, or a CDAE, is composed of a series of convolution and pooling layers. The units in the convolutional layers are organized in feature maps, where each feature map 
unit is connected to a local region of units in the (feature maps of the) previous layer, through a set of weights known as a filter. In this way the convolutional layer performs its function in the network, which is to detect locally connected features from the previous layer. This is an important role in spatial data such as images, where often local values are highly correlated.
For this reason, the number of weights to be learned is not dependent on the number of neurons, but on the number of feature maps and the size of the filter.

There are certain criteria for selecting an optimal filter size. It is important to note the effects of a small filter size versus a large one. If a network is required to detect/recognize an object which consists of a large number of pixels, a relatively large filter (e.g., $11 \times 11$ or $9 \times 9$), may be used. Otherwise, a smaller filter (e.g., $3 \times 3$ or $5 \times 5$) should be used.
In addition to feature capturing, the number of weights to be learned also depends on the filter size; larger filters can increase the training time of the network, and even prevent it from converging, due to too many parameters to be learned.
Regarding the intended gene expression based on processing the available ISH brain images, we can think of features distinguishing between genes as relatively small.

\begin{figure}[ht]
	\centering
	\includegraphics[width=\textwidth]{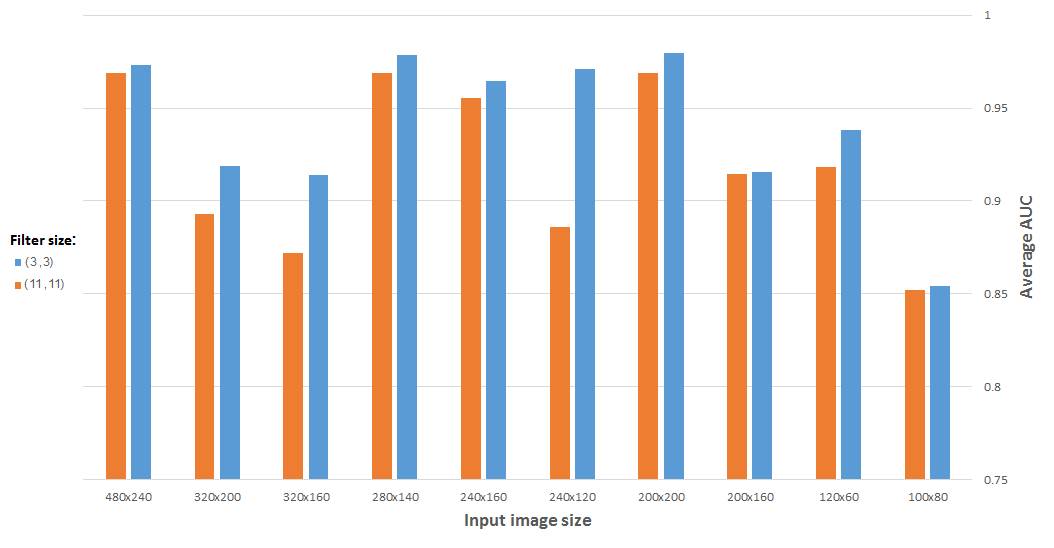}
	\caption{Average AUC versus filter size, using CDAE for different image sizes.}
	\label{fig:filter_size_all}
\end{figure}

Figure \ref{fig:filter_size_all} compares the best results obtained as a function of filter size, using a CDAE for images (and representation vectors) of different sizes. (All CDAEs have a fixed number of 8 feature maps for every convolutional layer.) As suspected, a $3 \times 3$ filter gives the best results for the problem in question.

\subsubsection{Representation Vector Size}

As discussed in Subsection \ref{sec:FuncISH}, \cite{Liscovitch2013FuncISH} presents FuncISH, a learning method of functional representations of ISH images, using a histogram of local descriptors on several scales. Their method captures compact representation vectors of size 2,004 out of the neural ISH images. Our aim is to find as small a representation as possible, that would still adequately represent the gene related to the ISH image, and classify it correctly to the GO categories.

\begin{figure}[ht!]
	\centering
	\includegraphics[width=\textwidth]{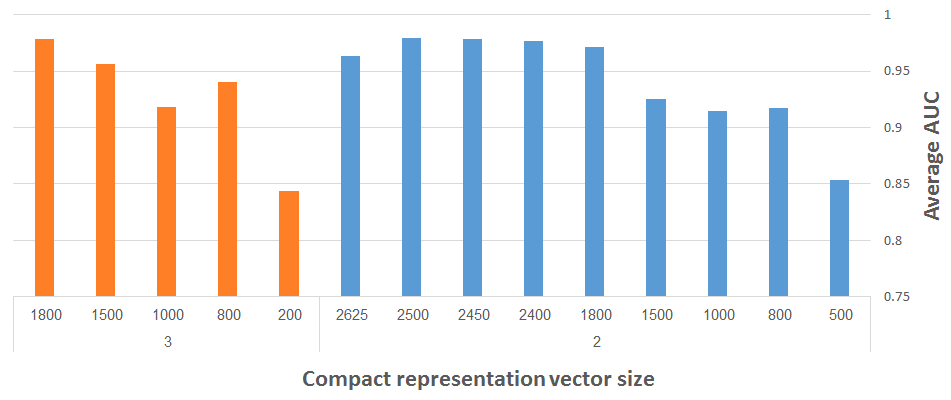}
	\caption{Maximum average AUC versus representation vector size, using CDAE with fixed-size ($3 \times 3$) filters; results on right and left hand side correspond to use of 2 and 3 max-pooling layers, respectively).}
	\label{fig:vector_length}
\end{figure}

Figure \ref{fig:vector_length} presents the maximum (over all runs) of the average AUC results for different representation vector sizes, using our CDAE. The plots depicted on the right- and left-hand side of the figure correspond to 2 and 3 max-pooling layers, respectively. (We used here merely 2 and 3 max-pooling layers on smaller input images in order to save considerable run-time during the experimentation. The performance of the system is reported eventually for the resampled images of size $960 \times 480$ using 4 max-pooling layers.) The filter size was fixed to $3 \times 3$, while other network hyper-parameters (e.g., learning rate, number of convolutional layers, etc.) were experimented with to find the best results. 
The following general observation can be made: The larger the representation vector, the better the performance. 
The combined objective should thus be to work with largest possible images and with as compact representation vectors as possible, while still meeting a required level of classification accuracy.

We now show how to improve the results of FuncISH \cite{Liscovitch2013FuncISH}, in conjunction with reducing the size of the representation vector below 2,004.

\subsection{Best AUC Results}
\label{sec:best-results}

The CDAE architecture for finding a compact representation for these down-sampled images is as follows:  \\
\big(1\big) \textbf{Input layer}: Consists of the raw image, resampled to $960 \times 480$  pixels, and corrupted by setting to zero 20\% of the pixels; \\
\big(2\big) Two sequential convolutional layers with four $3 \times 3$ filters each; \\
\big(3\big) Max-pooling layer of size $2 \times 2$; \\
\big(4\big) Two sequential convolutional layers with four $3 \times 3$ filters each; \\
\big(5\big) Max-pooling layer of size $2 \times 2$; \\
\big(6\big) Two sequential convolutional layers with four $3 \times 3$ filters each; \\
\big(7\big) Max-pooling layer of size $2 \times 2$; \\
\big(8\big) Two sequential convolutional layers with four $3 \times 3$ filters each; \\
\big(9\big) Max-pooling layer of size $2 \times 2$; \\
\big(10\big) One convolutional layers with four $3 \times 3$ filters each; \\
\big(11\big) One convolutional layers with a single $3 \times 3$ filters each; \\
\big(12\big) Unpooling layer of size $2 \times 2$; \\
\big(13\big) Two sequential deconvolutional layers with four $3 \times 3$ filters each; \\
\big(14\big) Unpooling layer of size $2 \times 2$; \\
\big(15\big) Two sequential deconvolutional layers with four $3 \times 3$ filters each; \\
\big(16\big) Unpooling layer of size $2 \times 2$; \\
\big(17\big) Two sequential deconvolutional layers with four $3 \times 3$ filters each; \\
\big(18\big) Unpooling layer of size $2 \times 2$; \\
\big(19\big) Two sequential deconvolutional layers with four $3 \times 3$ filters each; \\
\big(20\big) Deconvolutional layer with a single $3 \times 3$ filter, and \\
\big(21\big) \textbf{Output layer} with the uncorrupted resampled image.
\vspace*{0.5cm}

The network was initialized at random. In addition to the above parameters, we picked stride = 1. Also, the $\rm{ReLU}(.)$ activation function is used for all convolution and deconvolutional layers, except for the last deconvolutional layer, which uses $\rm{tanh}(.)$. The learning rate starts from 0.05 and is multiplied by 0.9 after each epoch, and the denoising effect is obtained by randomly removing 20\% of the pixels of every input image (i.e., 20\% of the neurons in the input layer), as was pointed out in Subsection~\ref{denoising_rate}. 
\pagebreak
\begin{figure}[ht!]
	\centering
	\includegraphics[height=.149\textheight, angle=90]{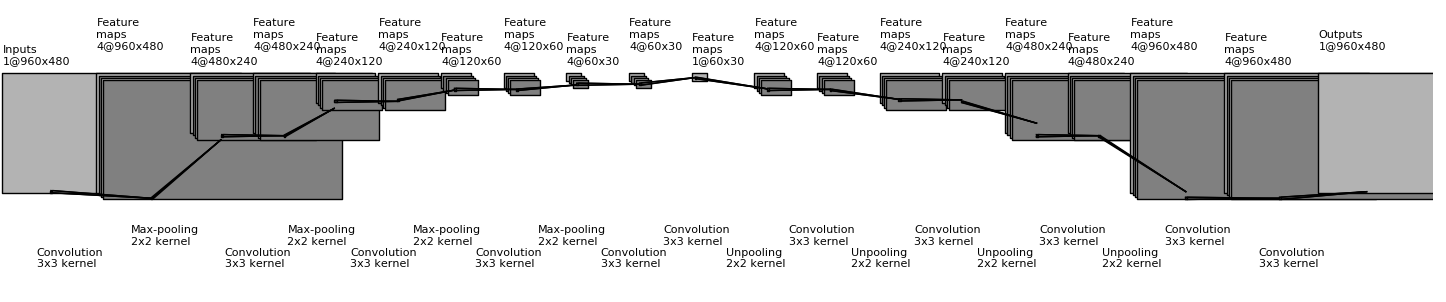}
	\caption{Illustration of our convolutional autoencoder, used to find a compact representation for the resampled images.}
	\label{fig:cae_first}
\end{figure}
\pagebreak

After training the CDAE, all layers past item 12 are removed, so that item 11 (the convolutional layer of size $1 \times 1,800$) becomes the output layer. Therefore, each image is mapped to a vector of 1,800 functional features.
Given a set of predefined GO annotations for each gene (where each GO category consists of 15--500 genes), we train a separate classifier for each biological category, similarly to~\cite{Liscovitch2013FuncISH}.
Specifically, we train an $L_2$-regularized logistic regression, using 5-fold cross-validation, as explained previously (see Subsection~\ref{sec:CDAE-classification}).

Our network yields remarkable AUC results for every category among the top 15 GO categories in~\cite{Liscovitch2013FuncISH}, i.e., the categories for which the best AUC scores were obtained by their FuncISH method, based on SIFT descriptor representations.
Using the compact representation extracted from our proposed CDAE architecture, coupled with the trained category classifiers, an exemplary AUC score of 1.0 was achieved for 13 of the 15 categories. The AUC values for the other two
categories was significantly better than those in~\cite{Liscovitch2013FuncISH} 
(0.999 vs. 0.87 and 0.96 vs. 0.86). While the average AUC score (of the top 15 categories) in \cite{Liscovitch2013FuncISH} is \textbf{0.92}, the average AUC for the same 15 GO categories using our CDAE is \textbf{0.997}, i.e., a \textbf{96\%} reduction in error. 

In addition, using this compact representation to classify all genes into the 2,081 GO categories yields an extraordinary average AUC result of \textbf{0.988}. Compared to that of~\cite{Zeng2015dcnn}, with an average AUC of $0.894 \pm 0.014$, we receive a \textbf{87\%} reduction in error.

What is also remarkable is that we were able not only to improve the accuracy of the classification, but to reduce the size of the compact representative vector by more than \textbf{10\%} relatively to \cite{Liscovitch2013FuncISH}, and by more than \textbf{80\%} with respect to the previous deep learning approach in \cite{Zeng2015dcnn}, i.e., to obtain a vector length of 1,800 from a vector lengths of 2,004 and 10,521, respectively.

Thus we succeeded in meeting the two main goals: (1) Improving the classification of GO categories using this representation, and (2) reducing the size of the representative vector.


\section{Concluding Remarks}
\label{sec:conclusion}

Many machine learning algorithms have been designed recently to predict GO annotations. For the task of learning meaningful functional representations of mammalian neural images, we employed in this paper deep learning techniques, and found convolutional denoising autoencoders to be very effective. Specifically, the presented scheme for feature learning of functional GO categories improved the previous state-of-the-art classification accuracy from an average AUC of 0.92 to 0.997, i.e., a 96\% reduction in error rate. In fact, we managed to classify 86\% of the categories with an exemplary AUC of 1. 

Furthermore, we demonstrated how to reduce the vector dimensionality by $10\%$ compared to the SIFT vectors used in \cite{Liscovitch2013FuncISH}, with very little degradation in accuracy.

Our results attest to the advantages of deep convolutional autoencoders, and especially the novel architecture as was applied here, for extracting meaningful information from very high resolution images and highly complex anatomical structures. Until gene product functions of all species are discovered, the use of CDAEs may well continue to serve in the ongoing design of novel biological experiments.

\bibliographystyle{plain}

\bibliography{gene-ontology}

\begin{thebibliography}{10}

\bibitem{Krizhevsky2009thesis}
Krizhevsky A.
\newblock Learning {M}ultiple {L}ayers of {F}eatures from {T}iny {I}mages.
\newblock {\em M.Sc. Thesis, Computer Science Department, University of
  Toronto}, 2009.

\bibitem{Krizhevsky2011DeepAE}
Krizhevsky A. and Hinton G.E.
\newblock Using very deep autoencoders for content-based image retrieval.
\newblock In {\em Proceedings of the European Symposium on Artificial Neural
  Networks}, 2011.

\bibitem{Krizhevsky2012ImageNet}
Krizhevsky A., Sutskever I., and Hinton G.E.
\newblock Image{N}et classification with deep convolutional neural networks.
\newblock In {\em Proceedings of the Advances in Neural Information Processing
  Systems}, pages 1106–--1114, 2012.

\bibitem{Perez2004annotation}
Perez A.J., Perez-Iratxeta C., Bork P., Thode G., and Andrade M.A.
\newblock Gene annotation from scientific literature using mappings between
  keyword systems.
\newblock {\em Bioinformatics}, 20(13):2084--2091, 2004.

\bibitem{Henry2012atlases}
Henry A.M. and Hohmann J.G.
\newblock High-resolution gene expression atlases for adult and developing
  mouse brain and spinal cord.
\newblock {\em Mammalian Genome}, 23:539--549, 2012.

\bibitem{GO2008project}
The Gene~Ontology Consortium.
\newblock The gene ontology project in 2008.
\newblock {\em Nucleic Acids Research}, 36:D440--D444, 2008.

\bibitem{Lowe2004recognition}
Lowe D. and Helmer S.
\newblock Object class recognition with many local features.
\newblock In {\em Proceedings of the Workshop on Generative Model Based
  Vision}, volume~12, pages 187--187, 2004.

\bibitem{Rumelhart1986errors}
Rumelhart D.E., Hinton G.E., and Williams R.J.
\newblock Learning representations by back-propagating errors.
\newblock {\em Nature}, 323:533–--536, 1986.

\bibitem{Lowe2004SIFT}
Lowe D.G.
\newblock Distinctive image features from scale-invariant keypoints.
\newblock {\em International Journal of Computer Vision}, 60(2):91--110, 2004.

\bibitem{Plessis2011why}
du~Plessis~L., Skunca N., and Dessimoz.
\newblock The what, where, how and why of gene ontology--a primer for
  bioinformaticians.
\newblock {\em Briefings in Bioinformatics}, 12(6):723--735, 2011.

\bibitem{Lein2007Genome}
Lein~E.S. et~al.
\newblock Genome-wide atlas of gene expression in the adult mouse brain.
\newblock {\em Nature}, 445:168--176, 2007.

\bibitem{Ng2009expression}
Ng~L. et~al.
\newblock An anatomic gene expression atlas of the adult mouse brain.
\newblock {\em Nature Neuroscience}, 12:356--362, 2009.

\bibitem{Davis2009Allen}
Davis F.P. and Eddy S.R.
\newblock A tool for identification of genes expressed in patterns of interest
  using the allen brain atlas.
\newblock {\em Bioinformatics}, 25:1647--1654, 2009.

\bibitem{Hinton2006dbn}
Hinton G.E., Osindero S., and Y.W. Teh.
\newblock A fast learning algorithm for deep belief nets.
\newblock {\em Neural Computation}, 18(7):1527--1554, 2006.

\bibitem{Lee2007sparse}
Lee H., Ekanadham C., and Ng~A.
\newblock Sparse deep belief net model for visual area v2.
\newblock In {\em Proceedings of the Advances in Neural Information Processing
  Systems}, volume~20, pages 873–--880, 2007.

\bibitem{Cohen2017deepbrain}
Cohen I., David E.O., Netanyahu N.S., Liscovitch N., and Chechik G.
\newblock Deep{B}rain: Functional representation of neural in-situ
  hybridization images for gene ontology classification using deep
  convolutional autoencoders.
\newblock In {\em Proceedings of the International Conference on Artificial
  Neural Networks}, number~2, pages 287–--296, 2017.

\bibitem{masci2011scae}
Masci J., Meier U., Ciresan D., and Schmidhuber J.
\newblock Stacked convolutional auto-encoders for hierarchical feature
  extraction.
\newblock In {\em International Conference on Artificial Neural Networks},
  pages 52--59, 2011.

\bibitem{Puniyani2013GINI}
Puniyani K. and Xing E.P.
\newblock {GINI}: From {ISH} images to gene interaction networks.
\newblock {\em PLOS Computational Biology}, 9:10, 2013.

\bibitem{Simonyan2014scale}
Simonyan K. and Zisserman A.
\newblock Very deep convolutional networks for large-scale image recognition.
\newblock {\em Computing Research Repository}, pages 1409--1556, 2014.

\bibitem{Ashburner2000GO}
Ashburner M., Ball C.A., Blake J.A., Botstein D., Butler H., and Cherry J.M.
\newblock Gene ontology: tool for the unification of biology.
\newblock {\em Nature Genetics}, 25(1):25--29, 2000.

\bibitem{Lowe2003panoramas}
Brown M. and Lowe D.G.
\newblock Recognising panoramas.
\newblock In {\em Proceedings of the 9th International Conference on Computer
  Vision}, pages 1218--1227, 2003.

\bibitem{Hawrylycz2011Multi}
Hawrylycz M., Ng~L., Page D., Morris J., Lau~C. Faber, S.~Faber V., Sunkin S.,
  Menon V., Lein E., and Jones A.
\newblock Multi-scale correlation structure of gene expression in the brain.
\newblock {\em Neural Networks}, 24:933--942, 2011.

\bibitem{Zitnik2014mold}
Zitnik M. and Zupan B.
\newblock Matrix factorization-based data fusion for gene function prediction
  in baker's yeast and slime mold.
\newblock In {\em Proceedings of the Pacific Symposium on Biocomputing}, pages
  400--411, 2014.

\bibitem{Zeiler2011adaptive}
Zeiler M.D., Taylor G.W., and Fergus R.
\newblock Adaptive deconvolutional networks for mid and high level feature
  learning.
\newblock In {\em Proceedings of the International Conference on Computer
  Vision}, pages 2018--2025, 2011.

\bibitem{zeiler2014Visualizing}
Zeiler M.D. and Fergus R.
\newblock Visualizing and understanding convolutional networks.
\newblock In {\em Proceedings of the European Conference on Computer Vision},
  pages 818--833, 2014.

\bibitem{Kordmahalleh2013Hierarchical}
Kordmahalleh M.M., Homaifar A., and Dukka B.k.C.
\newblock Hierarchical multi-label gene function prediction using adaptive
  mutation in crowding niching.
\newblock In {\em Proceedings of the IEEE International Conference on
  Bioinformatics and Bioengineering}, pages 1--6, 2013.

\bibitem{Liscovitch2013FuncISH}
Liscovitch N., Shalit U., and Chechik G.
\newblock Func{ISH}: {L}earning a functional representation of neural {ISH}
  images.
\newblock {\em Bioinformatics}, 29(13):i36--i43, 2013.

\bibitem{King2013patterns}
King O.D., Foulger R.E., Dwight S.S., White J.V., and Roth F.P.
\newblock Predicting gene function from patterns of annotation.
\newblock {\em Genome Research}, 13(5):896--904, 2013.

\bibitem{Pinoli2015Computational}
Pinoli P., Chicco D., and Masseroli M.
\newblock Computational algorithms to predict gene ontology annotations.
\newblock {\em BMC Bioinformatics}, 16(6):S4, 2015.

\bibitem{Vincent2010SDAE}
Vincent P., Larochelle H., Lajoie I., Bengio Y., and Manzagol P.
\newblock Stacked denoising autoencoders: Learning useful representations in a
  deep network with a local denoising criterion.
\newblock {\em The Journal of Machine Learning Research}, 11:3371--3408, 2010.

\bibitem{Vincent2008dae}
Vincent P., Larochelle H., Bengio Y., and Manzagol P.A.
\newblock Extracting and composing robust features with denoising autoencoders.
\newblock In {\em Proceedings of the 25th International Conference on Machine
  learning}, pages 1096--1103, 2008.

\bibitem{Werbos1974prediction}
Werbos P.J.
\newblock New {T}ools for {P}rediction and {A}nalysis in the {B}ehavioral
  {S}ciences.
\newblock {\em Ph.D. Thesis, Harvard University}, 1974.

\bibitem{Le2012building}
Le~Q.V., Monga R., Devin M., Chen K., Corrado G.S., Dean J., and Ng~A.
\newblock Building high-level features using large scale unsupervised learning.
\newblock In {\em Proceedings of the International Conference on Machine
  Learning}, 2012.

\bibitem{Behnke2003hierarchical}
Behnke S.
\newblock Hierarchical neural networks for image interpretation.
\newblock {\em Lecture Notes in Computer Science}, 2766:1--13, 2003.

\bibitem{Vembu2014prediction}
Vembu S. and Morris Q.
\newblock An efficient algorithm to integrate network and attribute data for
  gene function prediction.
\newblock In {\em Proceedings of the Pacific Symposium on Biocomputing}, pages
  388--399, 2014.

\bibitem{Sermanet2014overfeat}
Zhang X. Mathieu M. Fergus~R. Sermanet~P., Eigen~D. and LeCun Y.
\newblock Overfeat: Integrated recognition, localization and detection using
  convolutional networks.
\newblock In {\em Proceedings of the international conference on learning
  representations}, 2014.

\bibitem{Serre2005cortex}
Serre T., Wolf L., and Poggio T.
\newblock Object recognition with features inspired by visual cortex.
\newblock In {\em Proceedings of the Computer Vision and Pattern Recognition
  Conference}, pages 994--1000, 2005.

\bibitem{Zeng2015dcnn}
Zeng T., Li~R., Mukkamala R., Ye~J., and Ji~S.
\newblock Deep convolutional neural networks for annotating gene expression
  patterns in the mouse brain.
\newblock {\em BMC Bioinformatics}, (16):147, 2015.

\bibitem{Wolf2009cerebellar}
Wolf U., Rapoport M.J., and Schweizer T.A.
\newblock Evaluating the affective component of the cerebellar cognitive
  affective syndrome.
\newblock {\em Journal of Neuropsychology and Clinical Neuroscience},
  21(3):245--53, 2009.

\bibitem{Turchenko2015Caffe}
Turchenko V. and Luczak A.
\newblock Creation of a deep convolutional auto-encoder in {C}affe.
\newblock \url{https://arxiv.org/abs/1512.01596}, 2015.

\bibitem{LeCun1989handwritten}
LeCun Y., Boser B., Denker J.S., Henderson D., Howard R.E., Hubbard W., and
  Jackel L.D.
\newblock Backpropagation applied to handwritten zip code recognition.
\newblock {\em Neural Computation}, pages 541--551, 1989.

\bibitem{LeCun1998gradient}
LeCun Y., Bottou L., Bengio Y., and Haffner P.
\newblock Gradient-based learning applied to document recognition.
\newblock In {\em Proceedings of the IEEE}, volume~86, pages 2278–--2324,
  1998.

\end{thebibliography}

\end{document}